\documentclass{article}
\usepackage{ijcai25}

\usepackage{times}
\usepackage{soul}
\usepackage{url}
\usepackage[utf8]{inputenc}
\usepackage[small]{caption}
\usepackage{graphicx}
\usepackage{amsmath}
\usepackage{amsthm}
\usepackage{booktabs}
\usepackage{algorithm}
\usepackage{algorithmic}
\usepackage[switch]{lineno}
\usepackage{xcolor}
\usepackage{listings}
\usepackage[most]{tcolorbox}
\usepackage{graphics,graphicx,float,subcaption,booktabs,multirow,array,color,ifthen,colortbl,dblfloatfix,url,xparse,mathtools,patchcmd,amssymb,xspace,nicefrac,microtype,amsmath,amsfonts,bm,ragged2e,tikz,stackengine,etoolbox,xpatch,enumerate,xstring,tabularx,makecell,cuted,enumitem,tcolorbox,verbatim,multirow,footmisc,arydshln,duckuments,framed}

\definecolor{myorange}{HTML}{D79B00}
\definecolor{mypink}{HTML}{EA6B66}
\definecolor{myblue}{HTML}{7EA6E0}
\definecolor{myblue2}{rgb}{0.21,0.49,0.74}

\definecolor{ourpurple}{HTML}{7030A0} 
\definecolor{ouryellow}{HTML}{F4B402}

\definecolor{newGray}{rgb}{0.92, 0.92, 0.92}
\definecolor{NewBlue}{rgb}{0.95, 0.95, 1.0}

\usepackage[pagebackref,breaklinks,colorlinks,allcolors=myblue2]{hyperref}

\title{RoboHorizon: An LLM-Assisted Multi-View World Model for Long-Horizon Robotic Manipulation}

\author{
Zixuan Chen$^1$
\and
Jing Huo$^1$\and
Yangtao Chen$^1$\And
Yang Gao$^1$\\
\affiliations
$^1$State Key Laboratory for Novel Software Technology, Nanjing University, China\\
\emails
\{chenzx, huojing\}@nju.edu.cn,
502023330009@smail.nju.edu.cn,
gaoy@nju.edu.cn
}
\begin{document}

\maketitle

\begin{abstract}
    Efficient control in long-horizon robotic manipulation is challenging due to complex representation and policy learning requirements.
    Model-based visual reinforcement learning (RL) has shown great potential in addressing these challenges but still faces notable limitations, particularly in handling sparse rewards and complex visual features in long-horizon environments.
    To address these limitations, we propose the \textit{Recognize-Sense-Plan-Act} \textit{(RSPA)} pipeline for long-horizon tasks and further introduce \textbf{RoboHorizon}, an LLM-assisted multi-view world model tailored for long-horizon robotic manipulation. In RoboHorizon, pre-trained LLMs generate dense reward structures for multi-stage sub-tasks based on task language instructions, enabling robots to better \textit{recognize} long-horizon tasks. Keyframe discovery is then integrated into the multi-view masked autoencoder (MAE) architecture to enhance the robot's ability to \textit{sense} critical task sequences, strengthening its multi-stage perception of long-horizon processes. Leveraging these dense rewards and multi-view representations, a robotic world model is constructed to efficiently \textit{plan} long-horizon tasks, enabling the robot to reliably \textit{act} through RL algorithms.
    Experiments on two representative benchmarks, RLBench and FurnitureBench, show that RoboHorizon outperforms state-of-the-art visual model-based RL methods, achieving a 23.35\% improvement in task success rates on RLBench's 4 short-horizon tasks and a 29.23\% improvement on 6 long-horizon tasks from RLBench and 3 furniture assembly tasks from FurnitureBench.
\end{abstract}

\section{Introduction}

A general-purpose robotic manipulator for real-life applications should be capable of performing long-horizon tasks composed of multiple sub-task phases, such as kitchen organization or warehouse picking. For instance, kitchen organization requires a robot to complete tasks like sorting food items, placing them into the refrigerator, and cleaning the countertops, while warehouse picking might involve identifying orders, picking items, and packing them. But how can we design such a comprehensive robotic system? Traditionally, long-horizon robotic tasks are tackled using the "Sense-Plan-Act" (SPA) pipeline~\cite{DBLP:journals/automatica/Marton84,paul1981robot,murphy2019introduction}, which involves perceiving the environment, planning tasks based on a dynamic model, and executing actions through low-level controllers. A common approach to implementing this pipeline involves using visual and language encoders to extract task-relevant features for representation learning, followed by training control policies with model-based visual reinforcement learning (RL)~\cite{dalal2021accelerating,yamada2021motion,dalal2024plan}.
While the above solutions are somewhat effective, they still face significant challenges in complex long-horizon tasks: (1) Language and visual encoders struggle to capture the hierarchical structure and dependencies of multi-stage sub-tasks in long-horizon tasks; and (2) Environmental feedback in such tasks is often sparse, while RL policies heavily relies on the rational reward structure. The former limits the robot's ability to fully understand task dynamics and environmental context, while the latter further hinders the development of stable and effective long-horizon manipulation policies.
\begin{figure}[t!]
\centering
%\framebox[4.0in]{$\;$}
\includegraphics[width=0.5\textwidth]{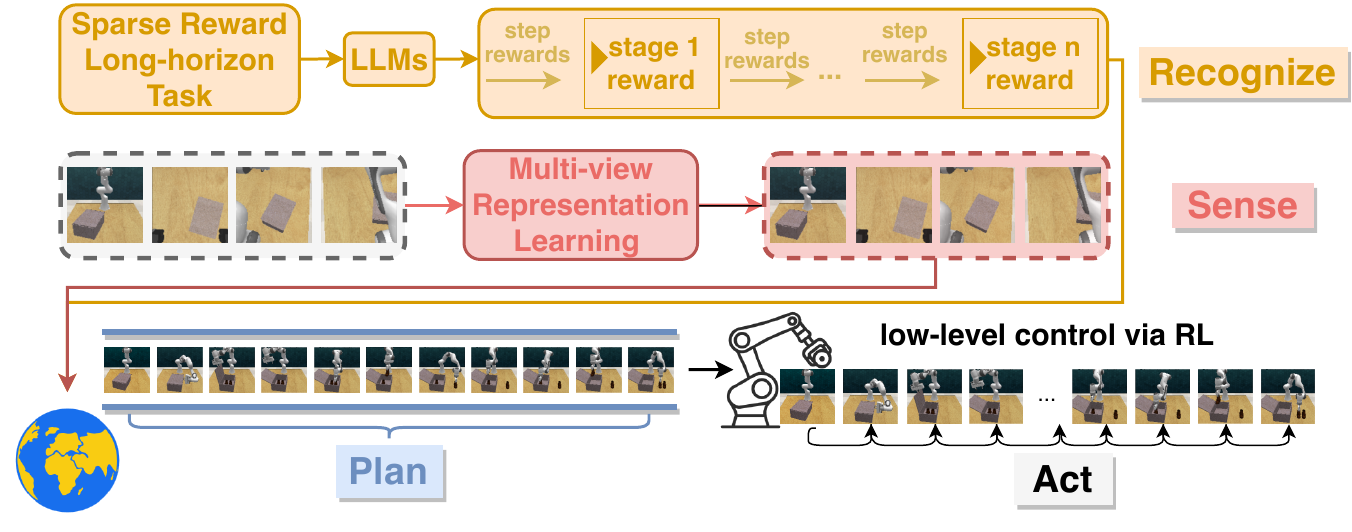} 
\caption{The proposed \textbf{\textit{RSPA pipeline}} for long-horizon robotic manipulation.}
\label{rspa}
\vspace{-0.4cm}
\end{figure}

\begin{figure*}[t!]
\centering
%\framebox[4.0in]{$\;$}
\includegraphics[width=0.85\textwidth]{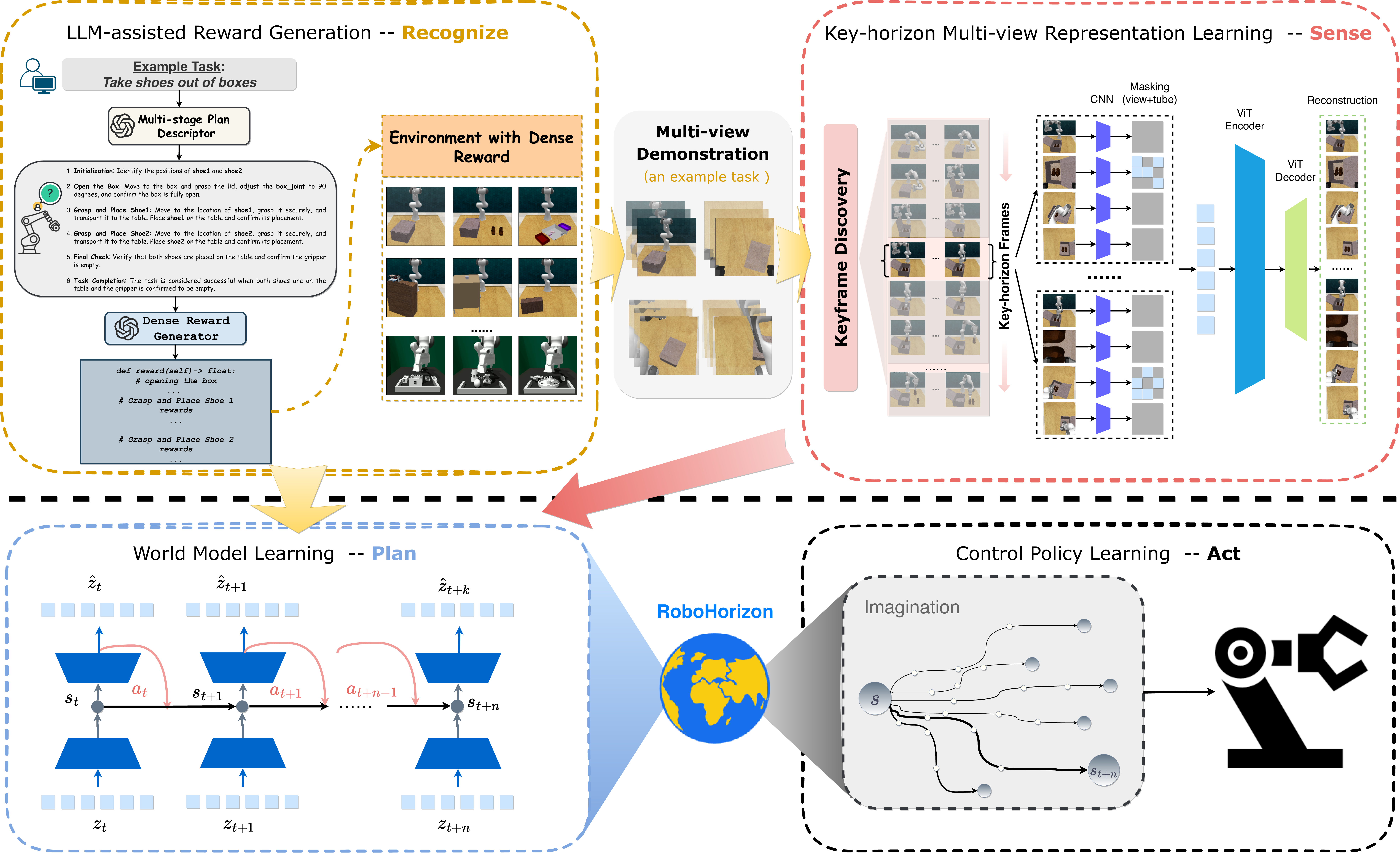} 
\caption{RoboHorizon overview, using the long-horizon robotic manipulation task \textit{``take shoes out of box"} in RLBench as the illustration example, following the proposed \textbf{\textit{RSPA pipeline}}.}
\label{RoboHorizon}
\vspace{-0.5cm}
\end{figure*}

Our key insight is that achieving stable execution of long-horizon tasks in model-based visual RL relies on \textbf{enabling robots to accurately understand tasks}, \textbf{perceive multi-stage interactions between the robot and objects in the environment}, and \textbf{learn stable control policies through a structured reward system}. How can robots be equipped with these capabilities? We propose leveraging pre-trained large language models (LLMs) and visual demonstrations captured by multi-view cameras to empower robots, primarily because:
\textbf{1)} LLMs have made significant advancements in robotics, demonstrating capabilities such as step-by-step planning~\cite{liang2023code,zeng2022socratic,ahn2022can,snell2022context}, goal-oriented dialogue~\cite{zeng2022socratic,ahn2022can,huang2022language}, sub-goals~\cite{huang2023voxposer,chen2024gravmad} and reward generation ~\cite{chiang2019learning,yu2023language} for robotic tasks based on language instructions.
and \textbf{2)} Observations from multi-camera views can significantly enhance a robot's visual manipulation capabilities, and this setup is becoming increasingly common in real-world applications. Execution trajectories captured from different viewpoints often share similar environmental dynamics and physical structures. Previous studies have explored learning control policies from multi-view offline data using model-based RL~\cite{seo2023multi} or imitation learning (IL)~\cite{goyal2023rvt,shridhar2023perceiver,ke20243d}.

Building on the above insight, we extend the traditional SPA pipeline into the \textit{Recognize-Sense-Plan-Act (RSPA)} pipeline, as illustrated in Fig.~\ref{rspa}, specifically designed to address the challenges in long-horizon robotic manipulation. 
At the core of this pipeline is \textbf{RoboHorizon}, an LLM-assisted multi-view world model that enables robots to execute tasks effectively in complex, long-horizon robotic scenarios.
To construct RoboHorizon, we first leverage pre-trained LLMs to generate a reasonable reward system for long-horizon tasks with sparse rewards. Unlike previous methods that use LLMs to generate reward signals directly applied to motion controllers~\cite{chiang2019learning,yu2023language}, which fail to address the complexities of multi-stage, long-horizon decision-making, our approach utilizes LLMs to divide long-horizon tasks into multi-stage sub-tasks. For each stage, dense stepwise rewards and intermediate rewards are generated, along with task reward code integrated into the environment interface, enabling robots to fundamentally \textit{\textbf{recognize}} the long-horizon tasks they need to execute.
Next, by collecting multi-view demonstrations of long-horizon manipulation tasks, we move beyond previous methods that directly perform representation learning on multi-view long-horizon demonstrations~\cite{seo2023multi,goyal2023rvt,shridhar2023perceiver}. Instead, we integrate the multi-view masked autoencoder (MAE) architecture~\cite{seo2023multi,he2021masked} with the keyframe discovery method in multi-stage sub-tasks~\cite{james2022q}, proposing a novel long-horizon key-horizon multi-view representation learning method. This method enables robots to more accurately \textit{\textbf{sense}} interactions between the robotic gripper and objects during critical multi-task stages.
Building on dense reward structures and the learned key-horizon multi-view representations, we construct RoboHorizon, enabling robots to effectively and efficiently \textit{\textbf{plan}} action trajectories for long-horizon tasks. Finally, we leverage imagined trajectories generated by the world model to train RL policies, enabling effective low-level \textit{\textbf{act}} capabilities for robots.
We evaluate RoboHorizon on 4 short-horizon and 6 long-horizon tasks from the RLBench~\cite{james2020rlbench} and 3 furniture assembly tasks from the FurnitureBench~\cite{heo2023furniturebench}, both representative testbeds for robotic manipulation under sparse rewards and multi-camera settings. RoboHorizon outperforms state-of-the-art model-based visual RL methods, with a 25.35\% improvement on short-horizon RLBench tasks and 29.23\% on long-horizon RLBench tasks and FurnitureBench assembly tasks. 

Our contributions can be summarized as follows:
\textbf{(1)} We introduce a novel \textit{Recognize-Sense-Plan-Act (RSPA)} pipeline for long-horizon robotic manipulation, which tightly integrates LLMs for dense reward structures generation (Recognize), key-horizon multi-view representation learning for interaction perceiving (Sense), a world model for action trajectories planning (Plan), and RL policies for robot control (Act).
\textbf{(2)} Based on the \textit{Recognize-Sense-Plan-Act (RSPA)} pipeline, we propose RoboHorizon, a robot world model specifically designed for long-horizon manipulation tasks, built upon LLM-generated dense reward structures and key-horizon multi-view representations. It enables efficient long-horizon task planning and ultimately ensures the stable execution of RL policies.
\textbf{(3)} We provide a comprehensive empirical study of RoboHorizon's performance in both short- and long-horizon manipulation tasks, validating RoboHorizon's effectiveness enabled by the proposed \textit{RSPA} pipeline.

\section{Related Work}
\paragraph{Methods for Long-horizon Manipulation Tasks}
Long-horizon robotic tasks are typically addressed through the ``Sense-Plan-Act" (SPA) pipeline~\cite{DBLP:journals/automatica/Marton84,paul1981robot,murphy2019introduction}. This pipeline involves comprehensive environment perception, task planning based on dynamic models of the environment, and action execution via low-level controllers. Traditional methods encompass a range of techniques, from operation planning~\cite{taylor1987sensor}, grasp analysis~\cite{miller2004graspit} to task and motion planning (TAMP)~\cite{garrett2021integrated} and skill-chaining~\cite{chenscar}. Recent approaches, on the other hand, integrate vision-driven learning techniques~\cite{mahler2016dex,sundermeyer2021contact}. These algorithms enable long-horizon decision-making in complex, high-dimensional action spaces~\cite{dalal2024plan}. However, they often face challenges in handling contact-rich interactions~\cite{mason2001mechanics,whitney2004mechanical}, are prone to cascading errors arising from imperfect state estimation~\cite{kaelbling2013integrated}, and require extensive manual engineering~\cite{garrett2020online}.
Our work leverages pre-trained large language models (LLMs) for task recognition, extending the traditional SPA pipeline into the \textit{Recognize-Sense-Plan-Act (RSPA)} pipeline, thereby significantly reducing the dependence on manual engineering. At the same time, our key-horizon multi-view representation learning method enhances the robot's ability to perceive contact-rich interactions. Together, these innovations effectively address cascading failures between sub-tasks, enabling robust long-horizon planning using the developed world model. 
% Our approach extends this pipeline to a recognize-sense-plan-act framework. By leveraging pre-trained large models for task recognition, we reduce the need for manual engineering. We enhance environmental perception through key-horizon representation learning from multi-view data, enabling robust long-horizon planning with a world model. Finally, reinforcement learning addresses contact-rich interactions, while online learning mitigates cascading failures between sub-tasks.

\paragraph{Visual Robotic Control with Multi-View Observation}
Building on the latest advancements in computer vision and robotics learning, numerous methods have been developed to leverage multi-view data from cameras for visual control~\cite{akinola2020learning,chen2021unsupervised,hsu2022vision,chen2023tild,shridhar2023perceiver,seo2023multi}. Some of these methods utilize self-supervised learning to obtain view-invariant representations~\cite{sermanet2018time}, learn 3D keypoints~\cite{chen2021unsupervised,shridhar2023perceiver,ke20243d}, or perform representation learning from different viewpoints~\cite{seo2023multi} to address subsequent manipulation tasks. However, these approaches are often limited to short-horizon robotic visual control tasks and lack the ability to handle long-horizon, multi-view robotic visual representations. In contrast, our work aims to develop a framework that learns key-horizon representations for multi-stage subtasks from long-horizon, multi-view visual demonstrations, enabling robots to tackle various complex long-horizon visual control tasks.
% \paragraph{World Models for Robotics}

\section{Method}
In this section, we detail the construction process of RoboHorizon, an LLM-assisted multi-view world model designed to achieve stable long-horizon robotic manipulation. 
First, we define the problem setting for the long-horizon manipulation tasks targeted in this work (Sec.~\ref{problem}). Next, we describe the LLM-assisted reward generation process (Recognize -- Sec.~\ref{recogniza}) and key-horizon multi-view representation learning method (Sense -- Sec.~\ref{sense}). Finally, we explain the development of the RoboHorizon world model (Plan -- Sec.~\ref{plan}) and the implementation of robot control through RL policies (Act -- Sec.~\ref{act}).

\subsection{Problem Setup}\label{problem}
We consider a long-horizon task as a Partially Observed Markov Decision Process (POMDP)~\cite{sutton1999policy} defined by $(\mathcal{S}, \mathcal{A}, \mathcal{T}, \mathcal{R}, p_0, \mathcal{O}, p_{O}, \gamma)$. Here, $\mathcal{S}$ represents the set of environment states, $\mathcal{A}$ the set of actions, $\mathcal{T}(s'|s,a)$ the transition probability distribution, $\mathcal{R}(s, a, s')$ the reward function, $p_0$ the initial state distribution, $\mathcal{O}$ the set of observations, $p_{O}(O|s)$ the observation distribution, and $\gamma$ the discount factor.
A sub-task $\omega$ is a smaller POMDP $(\mathcal{S}, \mathcal{A}_{\omega},\mathcal{T},\mathcal{R}_{\omega}, p_0^{\omega})$ derived from the full task's POMDP. For example, in the task ``\textit{take shoes out of the box}", the first sub-task is ``\textit{open the box}". The next sub-task, ``\textit{grasp and place shoe 1}", can only begin after the first is completed. When multiple sub-tasks are highly sequentially dependent, this forms the long-horizon tasks we focus on.
In our case, the observation space consists of all RGB images. The reward function is generated by large language models (LLMs), and the task description is provided to the agent in natural language. We also assume the availability of multi-view demonstration data for the task: $\zeta^v_{n} = \{o^v_0,\dots,o^v_n\}$, where $v \in \mathcal{V}$ represents available viewpoints, and $n$ is the total time step of the demonstrations.
\begin{figure}[t!]
\centering
%\framebox[4.0in]{$\;$}
\includegraphics[width=0.5\textwidth]{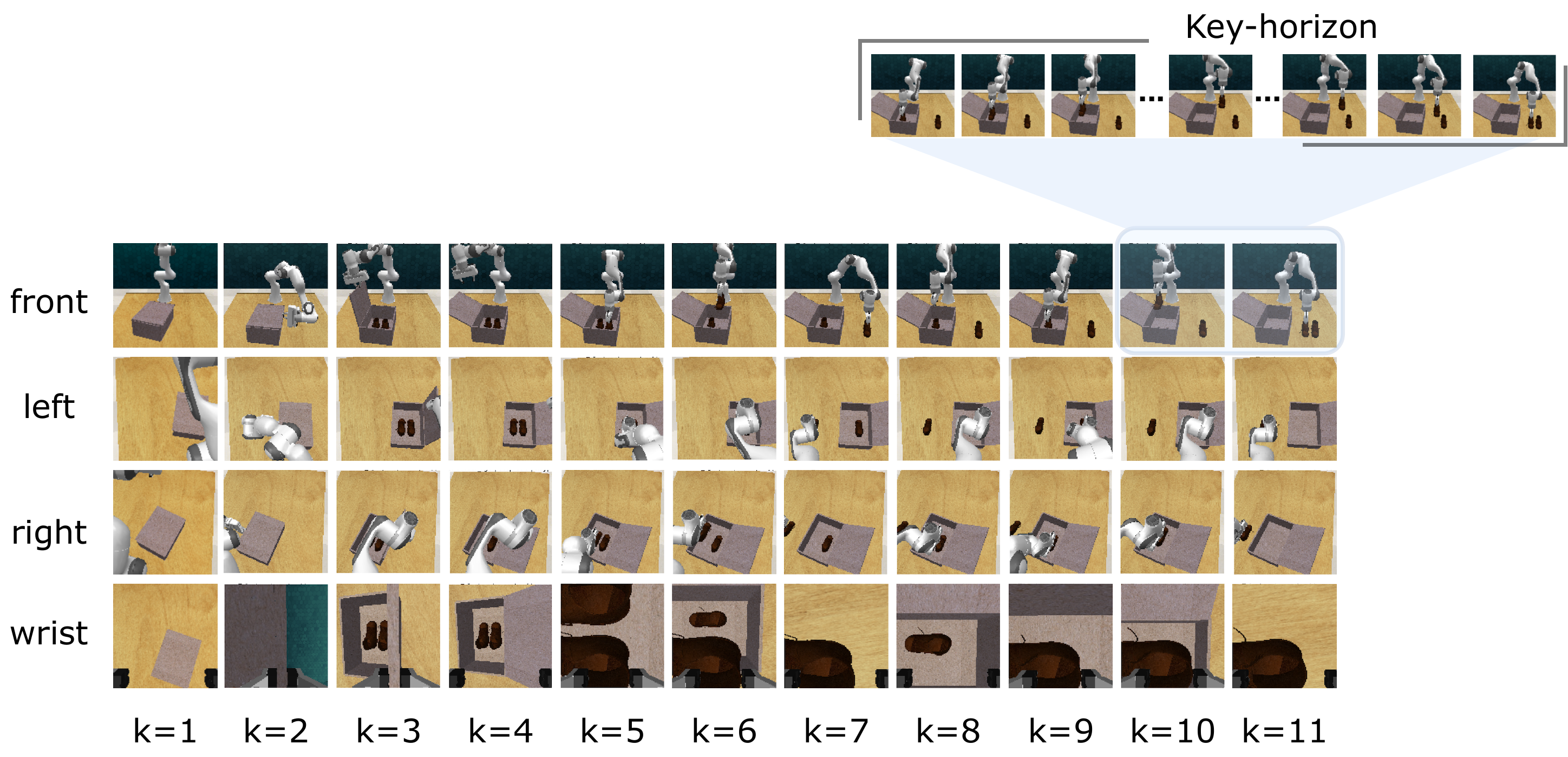} 
\caption{Visualizing the RGB observations of keyframes from four camera viewpoints for the \textit{take shoes out of the box} task using the keyframe discovery method, and displaying the key-horizon between the last two keyframes from the front viewpoint.}
\label{mv-keyframes}
\vspace{-0.5cm}
\end{figure}

\subsection{LLM-assisted Reward Generation -- Recognize}\label{recogniza}
We present the RoboHorizon overview in Fig.~\ref{rspa}, using the long-horizon manipulation task \textit{``take shoes out of box"} in RLBench as the illustration example.  
Specifically, given a task's language description, we prompt the LLMs to generate a corresponding task plan and encode dense rewards that align closely with each stage phase of the task. Following \citeauthor{yu2023language} (\citeyear{yu2023language}), we decompose the process of translating language into rewards into two stages: multi-stage tasks description and dense reward generation. It is worth noting that while we reference the architecture of \citeauthor{yu2023language} (\citeyear{yu2023language}), the internal prompts and tasks setting are entirely different.

% \begin{framed}
% \textbf{Stage 1} \\
% \small \textbf{Prompt:} Describe the long-horizon manipulation task using the following format: Graspable objects are shoe1 and shoe2. The box is successfully opened when the box\_joint is at 90 degrees. A shoe is grasped when the gripper holds either shoe1 or shoe2. A shoe is considered out of the box when detected on the table. The task is successful when both shoe1 and shoe2 are on the table.
% Remember: 1. Provide the motion description for the task; 2. Don’t include anything else.
% \\
% \small \textbf{Plan:} [Motion Description for the Task] 1. Initialize: Identify shoe1 and shoe2. 2. Open the Box: Move to the box. Adjust box\_joint to 90 degrees. Confirm the box is open. 3. Grasp and Place Shoe1: Move to shoe1. Grasp shoe1. Move to the table. Place shoe1 on the table. Confirm shoe1 is on the table. 4. Grasp and Place Shoe2: Move to shoe2. Grasp shoe2. Move to the table. Place shoe2 on the table. Confirm shoe2 is on the table. 5. Final Check: Ensure both shoes are on the table. Confirm the gripper is empty. 6. Task Completion: The task is successful when both shoes are on the table and the gripper is empty.
% \end{framed}
As can be seen from the \textbf{\textcolor{myorange}{Recognize}} part in Fig.~\ref{rspa}, in the Stage 1, we employ a pre-trained LLM as the Multi-stage Plan Descriptor that interprets and expands user input into detailed language descriptions of the required robot motions using predefined templates. To enable the multi-stage plan descriptor to generate a coherent structure for long-horizon tasks, we create a prompt template that outlines the current robot task setting. This leverages the pre-trained LLM's internal knowledge of motion planning to produce detailed motion descriptions. 
In the Stage 2, we deploy another LLM as the Dense Reward Generator to convert these motion descriptions into corresponding reward functions. We approach this as a coding task, leveraging the pre-trained LLM's understanding of coding and code structure. Four types of prompts guide the dense reward generator in generating the reward codes: i) task stage descriptions based on the task environment interface, ii) examples of expected reward generator responses, iii) constraints and rules for the reward encoder, and iv) specific task descriptions. 
Due to space limitations, example prompts for Stage 1 and Stage 2 of the \textit{take shoes out of box} task are in the Appendix. Additionally, while any pre-trained language model can be used for reward generation, we find that only GPT-4o (OpenAI, 2024) reliably generates correct plans and rewards for all tasks. A detailed data flow of LLM-assisted reward generation is in the Appendix.

\subsection{Key-horizon Multi-view Representation Learning -- Sense}\label{sense}
To enable robots to learn multi-stage interaction representations from long-horizon multi-view visual demonstrations, we propose the Key-Horizon Multi-View Masked Autoencoder (KMV-MAE) based on the MV-MAE architecture~\cite{seo2023multi}.
As shown in the \textbf{\textcolor{mypink}{Sense}} part of Fig.~\ref{rspa}, our KMV-MAE method extracts key-horizons from multi-view demonstrations using the keyframe discovery method~\cite{james2022q}. We then perform view-masked training on these key-horizons and use a video masked autoencoder to reconstruct missing pixels from masked viewpoints. Following prior work~\cite{seo2023masked,seo2023multi}, we mask convolutional features instead of pixel patches and predict rewards to capture fine-grained details essential for long-horizon visual control.

\paragraph{Keyframe Discovery.} The keyframe discovery method in our KMV-MAE, following previous works~\cite{james2022q,goyal2023rvt,shridhar2023perceiver,ke20243d}, identifies keyframes based on near-zero joint velocities and unchanged gripper states. As illustrated in Fig.~\ref{mv-keyframes}, this method captures each viewpoint's keyframe $\mathcal{K}^v = \{k^v_1, k^v_2, \ldots, k^v_m\}$ in the \textit{take shoes out of the box} task from the demonstration $\zeta^v$, where $k$ represents the keyframe number. The corresponding time steps in the demonstration for each keyframe are $\{t_{k_1},\dots,t_{k_m}\}$. 
Each adjacent keyframe pair $k^v_i$ and $k^v_{i+1}$ then forms a key-horizon $h_i = \{o^v_{t_{k_i}},\dots,o^v_{t_{k_{i+1}}}\}$. Notably, the number of RGB observations in each key-horizon varies, depending on the time step difference between the adjacent keyframes in the demonstration.

\paragraph{View\&Tube Masking and Reconstruction.} To extract more interaction information from multi-view long-horizon demonstrations, we propose a view\&tube masking method. For each frame, we randomly mask all features from three of the four viewpoints, while the remaining viewpoint has 95\% of its patches randomly masked. Across the key-horizon, the unmasked viewpoint follows the tube masking strategy~\cite{tong2022videomae}. This approach enhances cross-view feature learning, accounts for temporal correlations within a single viewpoint, reduces information leakage, and improves temporal feature representation.  
We integrate video masked autoencoding~\cite{feichtenhofer2022masked,tong2022videomae} with the view\&tube masking operation. Vision Transformer (ViT)~\cite{dosovitskiy2020image} layers encode unmasked feature sequences across all viewpoints and frames. Following \citeauthor{seo2023masked} (\citeyear{seo2023masked,seo2023multi}), we concatenate mask tokens with the encoded features and add learnable parameters for each viewpoint and frame to align features with mask tokens. Finally, ViT layers decode the features, projecting them to reconstruct pixel patches while also predicting rewards to encode task-relevant information. This representation learning process can be summarized as:

% Let $o^{v}_{t,T}= \{o^{v}_{t}, ..., o^{v}_{t+T-1}\}$ be a video from a viewpoint $v \in \mathcal{V}$ where $t$ is current timestep, $T$ is window size of the video autoencoder, and
% $\mathcal{V}$ is a set of available viewpoints. 
Given demonstration videos $\zeta^v_{n} = \{o^v_0,\dots,o^v_n\}$, after $m$ keyframes $\{k^v_1, k^v_2, \ldots, k^v_m\}$ are extracted by the keyframe discovery method, they become in the form of containing $m-1$ key-horizon: $\zeta^v = \{h^{v}_1,\dots,h^{v}_{m-1}\}_{v \in \mathcal{V}}$ from multiple viewpoints. Given LLM-assisted generated rewards $r = \{r_{1}, ..., r_{n}\}$, and a mask ratio of $m$, KMV-MAE consists of following components:
\begin{align}
&\text{Convolution stem:} &&l^{v}_{i}= f^{\tt{conv}}_{\phi}(h^{v}_{i})\nonumber \\
&\text{View\&Tube masking:} &&l_{i}^{m}\sim p^{\tt{mask}}(l_{i}^{m}\,|\,\{h^{v}_{i}\}_{v \in \mathcal{V}}, m)
\nonumber \\
&\text{ViT encoder:} &&z_{i}^{m}\sim p_{\phi}(z_{i}^{m} \,|\,l_{i}^{m}) \label{eq:mvmae}\\
&\text{ViT decoder:} 
&&\begin{aligned}
\begin{cases}
&\{\hat{h}^{v}_{i}\}_{v \in \mathcal{V}}\sim p_\phi(\{\hat{h}^{v}_{i}\}_{v \in \mathcal{V}}\,|\,z_{i}^{m})\\
&\hat{r}_{t_{k_i},t_{k_{i+1}}}\sim p_{\phi}(\hat{r}_{t_{k_i},t_{k_{i+1}}} \,|\,z_{i}^{m})
\end{cases}
\end{aligned} \nonumber
\end{align}
Finally the model is trained to reconstruct key-horizon pixels and predict rewards, i.e., minimizing the negative log-likelihood to optimize the model parameter $\phi$ as follows:
\begin{align}
    &\mathcal{L}^{\tt{kmvmae}}(\phi) = -\ln p_{\phi}(\{h^{v}_{i}\}_{v \in \mathcal{V}}\,|\,z^{m}_{i}) -\ln p_{\phi}(r_{t_{k_i},t_{k_{i+1}}}\,|\,z^{m}_{i}) \nonumber
\end{align}

\subsection{RoboHorizon World Model -- Plan}\label{plan} 
For the \textbf{\textcolor{myblue}{Plan}} part in Fig.~\ref{rspa}, we construct RoboHorizon following previous works\cite{seo2023masked,seo2023multi}, implementing it as a variant of the Recurrent State Space Model (RSSM)~\cite{hafner2019learning}. The model uses frozen autoencoder representations from the previous key-horizon multi-view representation learning as inputs and reconstruction targets. RoboHorizon includes the following components:
\begin{gather}
\begin{aligned}
&\text{Encoder:} &&s_t\sim q_\theta(s_{t} \,|\,s_{t-1},a_{t-1}, z_{t}) \\
&\text{Decoder:} &&\begin{aligned} \begin{cases} 
&\hat{z}_t\sim p_\theta(\hat{z}_{t} \,|\,s_{t})\\
&\hat{r}_t\sim p_\theta(\hat{r}_{t} \,|\,s_{t})
\end{cases}
\end{aligned} \\
&\text{Dynamics model:} &&\hat{s}_t\sim p_\theta(\hat{s}_{t} \,|\,s_{t-1}, a_{t-1})
\label{eq:kmvmwm}
\end{aligned}
\end{gather}
The encoder extracts state $s_{t}$ from the previous state $s_{t-1}$, previous action $a_{t-1}$, and current autoencoder representations $z_{t}$. The dynamics model predicts $s_{t}$ without access to $z_{t}$, allowing forward prediction. 
% Following \cite{hafner2020mastering}, we use discrete latents for $s_{t}$. 
The decoder reconstructs $z_{t}$ to provide learning signals for model states and predicts $r_{t}$ to compute rewards from future states without decoding future autoencoder representations. All model parameters $\theta$ are optimized jointly by minimizing the negative variational lower bound~\cite{kingma2013auto}.:
\begin{gather}
\begin{aligned}
    &\mathcal{L}^{\tt{wm}}(\theta) = -\ln p_\theta(z_{t} \,|\,s_{t})-\ln p_\theta(r_{t} \,|\,s_{t}) \\
    &\quad+  \beta\,\text{KL}\big[ q_{\theta}(s_{t}|s_{t-1},a_{t-1},z_{t}) \,\Vert\,  p_{\theta}(\hat{s}_{t}|s_{t-1},a_{t-1}) \big], \nonumber
\end{aligned}
\end{gather}
where $\beta$ is a scale hyperparameter.

\subsection{Control Policy Learning -- Act}\label{act}
For the \textbf{Act} part in Fig.~\ref{rspa}, we build on the approach of \cite{seo2023masked,seo2023multi} and adopt the actor-critic framework from DreamerV2~\cite{hafner2020mastering}. The goal is to train a policy that maximizes predicted future values by backpropagating gradients through the RoboHorizon world model. Specifically, we define a stochastic actor and a deterministic critic as:
\[
\text{Actor:} \quad \hat{a}_{t} \sim p_{\psi}(\hat{a}_{t}\,|\,\hat{s}_{t}) \quad \text{Critic:} \quad v_{\xi}(\hat{s}_{t}) \approx \mathbb{E}_{p_{\theta}}\left[\sum_{i \leq t}\gamma^{i - t} \hat{r}_{i}\right]
\]
Here, the sequence $\{(\hat{s}_{t}, \hat{a}_{t}, \hat{r}_{t})\}_{t=1}^{H}$ is predicted from the initial state $\hat{s}_{0}$ using the stochastic actor and dynamics model from Eq.~\ref{eq:kmvmwm}. Unlike previous work, we set $H$ to match the length of each key-horizon in the long-horizon task, with each key-horizon sequence having a different duration.
Given the $\lambda$-return~\cite{schulman2015high} defined as:
\[
V_{t}^{\lambda} \doteq \hat{r}_{t} + \gamma
\begin{cases}
(1 - \lambda)v_{\xi}(\hat{s}_{t+1}) + \lambda V_{t+1}^{\lambda} & \text{if } t<H \\
v_{\xi}(\hat{s}_{H}) & \text{if } t=H
\end{cases}
\label{eq:lambda_return}
\]
the critic is trained to regress the $\lambda$-return, while the actor is trained to maximize the $\lambda$-return with gradients backpropagated through the world model. 
To enable the robot to execute long-horizon tasks more reliably, we introduce an auxiliary behavior cloning loss that encourages the agent to learn expert actions while interacting with the environment. To achieve this, we follow the setup of \cite{james2022q,seo2023multi} to acquire the demonstration. Specifically, at each time step, given an expert action $a_{t}^{\tt{e}}$, the objective of auxiliary behavior cloning is $\mathcal{L}^{\tt{BC}} = -\ln p_{\psi}(a_{t}^{\tt{e}}|s_{t})$.
Thus, the objective for the actor network and the critic network is:
\begin{align}
    &\mathcal{L}^{\tt{critic}}(\xi)\doteq\mathbb{E}_{p_{\theta},p_{\psi}}\left[\sum^{H-1}_{t=1} \frac{1}{2} \left(v_{\xi}(\hat{s}_{t}) - \text{sg}(V_{t}^{\lambda})\right)^{2}\right] \nonumber \\
    &\mathcal{L}^{\tt{actor}}_{\tt{BC}}(\psi)\doteq \mathbb{E}_{p_{\theta},p_{\psi}} \left[-V_{t}^{\lambda} - \eta\,\text{H}\left[a_{t}|\hat{s}_{t}\right] \right] + \mathcal{L}^{\tt{BC}} \nonumber
\end{align}
where sg is a stop gradient operation, $\eta$ is a scale hyperparameter for an entropy $\text{H}\left[a_{t}|\hat{s}_{t}\right]$.
Thus, with the generated dense reward structure, the training objective for the \textbf{sense}, \textbf{plan}, and \textbf{act} processes in RoboHorizon is to minimize the following objective function:
\begin{equation}
    \mathcal{L}^{\tt{RoboHorizon}} = \underbrace{\mathcal{L}^{\tt{kmvmae}}}_{\text{\textbf{\textcolor{mypink}{Sense}}}} + \underbrace{\mathcal{L}^{\tt{wm}}}_{\text{\textbf{\textcolor{myblue}{Plan}}}} + \underbrace{\mathcal{L}^{\tt{critic}} + \mathcal{L}^{\tt{actor}}_{\tt{BC}}}_{\text{\textbf{Act}}}
\end{equation}

\begin{table*}[t]
\caption{Success Rates (\%) on \textcolor{ouryellow}{4 short-horizon tasks} (\colorbox{newGray}{in RLBench}) and \textcolor{ourpurple}{9 long-horizon tasks} (\colorbox{newGray}{\textcolor{ourpurple}{6 in RLBench}}, \colorbox{NewBlue}{\textcolor{ourpurple}{3 in FurnitureBench}}). Results are averaged over 5 seeds.}
\label{tab:main_result}
\centering
\fontsize{7.5}{8.5}\selectfont % Slightly larger font size for readability
\setlength{\tabcolsep}{3pt} % Moderate column spacing
\renewcommand{\arraystretch}{1.2} % Balanced row spacing
\begin{tabular}{@{}c|cccc|cccccc|ccc|cc@{}}
\toprule
\multirow{2}{*}{\textbf{Model}} & \multicolumn{4}{c|}{\textcolor{ouryellow}{\cellcolor{newGray}\textbf{Short-Horizon Tasks}}} & \multicolumn{6}{c|}{\textcolor{ourpurple}{\cellcolor{newGray}\textbf{Long-Horizon Tasks (RLBench)}}} & \multicolumn{3}{c|}{\textcolor{ourpurple}{\cellcolor{NewBlue}\textbf{FurnitureBench}}} & \multicolumn{2}{c}{\textbf{Average}} \\
 & \thead{\fontsize{7}{7.5}\selectfont Phone \\ \fontsize{7}{7.5}\selectfont on Base} 
 & \thead{\fontsize{7}{7.5}\selectfont Take \\ \fontsize{7}{7.5}\selectfont Umbrella} 
 & \thead{\fontsize{7}{7.5}\selectfont Put Rubbish \\ \fontsize{7}{7.5}\selectfont in Bin} 
 & \thead{\fontsize{7}{7.5}\selectfont Stack \\ \fontsize{7}{7.5}\selectfont Wine} 
 & \thead{\fontsize{7}{7.5}\selectfont Take \\ \fontsize{7}{7.5}\selectfont Shoes} 
 & \thead{\fontsize{7}{7.5}\selectfont Put \\ \fontsize{7}{7.5}\selectfont Shoes} 
 & \thead{\fontsize{7}{7.5}\selectfont Empty \\ \fontsize{7}{7.5}\selectfont Container} 
 & \thead{\fontsize{7}{7.5}\selectfont Put \\ \fontsize{7}{7.5}\selectfont Books} 
 & \thead{\fontsize{7}{7.5}\selectfont Put \\ \fontsize{7}{7.5}\selectfont Item} 
 & \thead{\fontsize{7}{7.5}\selectfont Slide Cabinet \\ \fontsize{7}{7.5}\selectfont \& Place Cups} 
 & \thead{\fontsize{7}{7.5}\selectfont Cabinet} 
 & \thead{\fontsize{7}{7.5}\selectfont Lamp} 
 & \thead{\fontsize{7}{7.5}\selectfont Round \\ \fontsize{7}{7.5}\selectfont Table} 
 & \textbf{\fontsize{7}{7.5}\selectfont Short Avg.} 
 & \textbf{\fontsize{7}{7.5}\selectfont Long Avg.} \\
\midrule
TCN+WM   & 5.2 & 4.1 & 2.4 & 2.2 & 0   & 0   & 0   & 0.4 & 0   & 0   & 1.0 & 6.5 & 8.2 & 3.48 & 1.79 \\
CLIP+WM  & 13.3 & 15.7 & 12.2 & 11.8 & 0   & 0   & 0   & 2.2 & 0   & 0   & 3.8 & 11.7 & 15.5 & 13.25 & 3.69 \\
MAE+WM   & 20.4 & 19.1 & 18.6 & 19.6 & 0   & 0   & 0   & 2.5 & 0   & 0   & 8.5 & 16.3 & 20.9 & 19.43 & 5.36 \\
MWM      & 32.5 & 30.8 & 28.4 & 29.7 & 1.2 & 0.8 & 2.1 & 3.3 & 1.1 & 0.5 & 15.2 & 27.5 & 32.0 & 30.35 & 9.30 \\
MV-MWM   & 52.6 & 50.3 & 48.9 & 49.1 & 3.5 & 2.7 & 4.6 & 10.4 & 5.2 & 2.9 & 26.6 & 43.5 & 46.7 & 50.23 & 16.23 \\
\midrule
\textbf{RoboHorizon} & \textbf{78.4} & \textbf{75.2} & \textbf{74.8} & \textbf{73.9} & \textbf{36.5} & \textbf{31.2} & \textbf{40.5} & \textbf{58.4} & \textbf{48.2} & \textbf{33.6} & \textbf{41.0} & \textbf{58.5} & \textbf{61.3} & \textbf{75.58} {\fontsize{6.5}{6.5}\selectfont\textcolor{green}{(25.35\% $\uparrow$)}} & \textbf{45.47} {\fontsize{6.5}{6.5}\selectfont\textcolor{green}{(29.23\% $\uparrow$)}} \\
\bottomrule
\end{tabular}
\vspace{-0.5cm}
\end{table*}

\section{Experiments}
We design our experiments to explore the following questions: (i) How does RoboHorizon, following the \textit{RSPA} pipeline, perform compared to SPA-driven model-based reinforcement learning (RL) baselines in short- and long-horizon manipulation tasks? (ii) If SPA-based baseline methods also adopt stepwise rewards generated by LLMs but do not aware the staged reward for each sub-task stage, does RoboHorizon still maintain its advantage? (iii) To what extent do RoboHorizon's key design components impact its overall performance?

\paragraph{Environmental Setup}
For quantitative evaluation, we adopt a demonstration-driven RL setup to address visual robotic manipulation tasks in RLBench~\cite{james2020rlbench} and FurnitureBench~\cite{heo2023furniturebench}. In both benchmarks, we rely on limited environment interactions and expert demonstrations. All experiments use only RGB observations from each camera, without incorporating proprioceptive state or depth information.  
Following previous studies \cite{james2022q,seo2023multi}, we populate the replay buffer with expert demonstrations, and the RL agent outputs relative changes in the gripper's position. For all tasks, 50 expert demonstrations are provided for each camera view. For FurnitureBench, we use a low-randomness environment initialization setup. Due to space limitations, more detailed experimental setups are provided in the Appendix.

\begin{figure}[t!]
    \centering
    \includegraphics[width=0.5\textwidth]{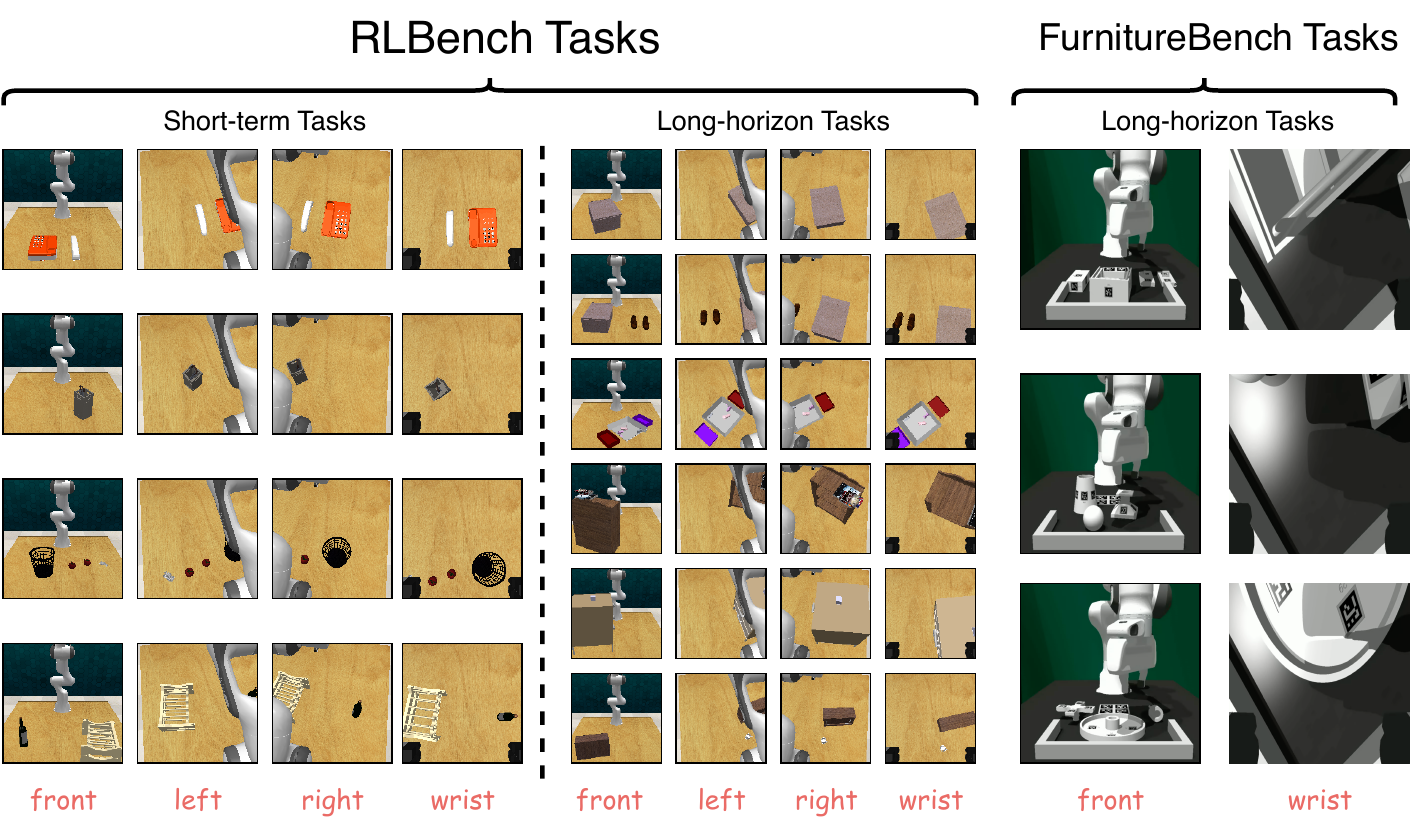}
    \caption{Visualization of multi-view demonstrations from front, left, right, and wrist cameras for 10 RLBench tasks, and from front and wrist cameras for 3 FurnitureBench tasks.}
    \label{fig:task}
    \vspace{-0.5cm}
\end{figure}

\paragraph{Multi-view Camera Setup}
We adopt a multi-view observation and single-view control approach \cite{seo2023multi}, suitable for scenarios where multiple cameras are available during training, but the robot relies on a single camera during deployment. For RLBench tasks, we use multi-view data from front, wrist, left, and right cameras to enhance the robot's perception of long-horizon tasks and the environment, while training a RL agent that operates solely on front camera input. For FurnitureBench tasks, we use multi-view data from front and wrist cameras with the same training and control setup.
We conduct experiments on 10 representative tasks in RLBench, including 4 short-horizon tasks (\textit{phone on base, take umbrella out of stand, put rubbish in bin, stack wine}) and 6 long-horizon tasks (\textit{take shoes out of box, put shoes in box, empty container, put books on bookshelf, put item in drawer, slide cabinet open and place cups}), as well as 3 long-horizon furniture assembly tasks in FurnitureBench (\textit{cabinet, lamp, round table}), as shown in Fig.~\ref{fig:task}. In these tasks, the front, left, right cameras provide a wide view of the robot’s workspace, while the wrist camera offers close-up views of the target objects.

\paragraph{Baselines}
To compare with SPA-driven model-based RL methods using manually defined rewards, we select MV-MWM~\cite{seo2023multi} and MWM~\cite{seo2023masked} as baselines. The former lacks key-horizon representation learning, while the latter lacks multi-view key-horizon representation learning. Both baselines rely on manually defined rewards and the same amount of training data. Additionally, we adopt various representation learning methods to train world models, further demonstrating the effectiveness of our key-horizon multi-view representation learning in long-horizon tasks. The comparison methods include CLIP+WM~\cite{radford2021learning}, MAE+WM~\cite{he2021masked}, and TCN+WM~\cite{sermanet2018time}. Specifically, RoboHorizon, MV-MWM, MWM, and TCN+WM learn representations from scratch, whereas CLIP+WM and MAE+WM use frozen pre-trained representations.  
More details about the experimental baselines are provided in the Appendix.

\subsection{Performance Comparison}
In this section, we conduct experiments on 4 short-horizon and 6 long-horizon tasks from RLBench, as well as 3 furniture assembly tasks from FurnitureBench, to address the three initial questions.

\paragraph{SPA-driven model-based RL baselines vs. RoboHorizon}
Table~\ref{tab:main_result} compares the success rates of RoboHorizon and five SPA-driven baselines on \textcolor{ouryellow}{4 short-horizon tasks} (in RLBench) and \textcolor{ourpurple}{9 long-horizon tasks} (6 in RLBench, 3 in FurnitureBench). RoboHorizon outperforms all baseline methods, achieving the highest average success rate across all tasks. Specifically, it exceeds MV-MWM by 25.35\% on the 4 short-horizon tasks and by 29.23\% on the 9 long-horizon tasks. This result shows that with LLM-assisted reward structures and key-horizon multi-view representation learning, RoboHorizon excels in both short-horizon pick-and-place tasks and long-horizon tasks with multiple sub-tasks and numerous target objects, achieving more stable manipulation. While MV-MWM benefits from multi-view representation learning, outperforming MWM, MAE+WM, CLIP+WM, and TCN+WM in representation, it still struggles with long-horizon tasks. Additionally, the manually designed reward structures are not robust, leading to inconsistent performance across all tasks for MV-MWM, MWM, MAE+WM, CLIP+WM, and TCN+WM.

These  experimental results effectively answer our first question:\textbf{ RoboHorizon outperforms SPA-driven baselines with manually defined rewards in both short- and long-horizon tasks, with a particularly strong advantage in long-horizon scenarios. }This outcome strongly validates the capability of our proposed \textit{RSPA} pipeline in handling long-horizon tasks.

\begin{figure*}[t!]
    \centering
    \includegraphics[width=0.95\textwidth]{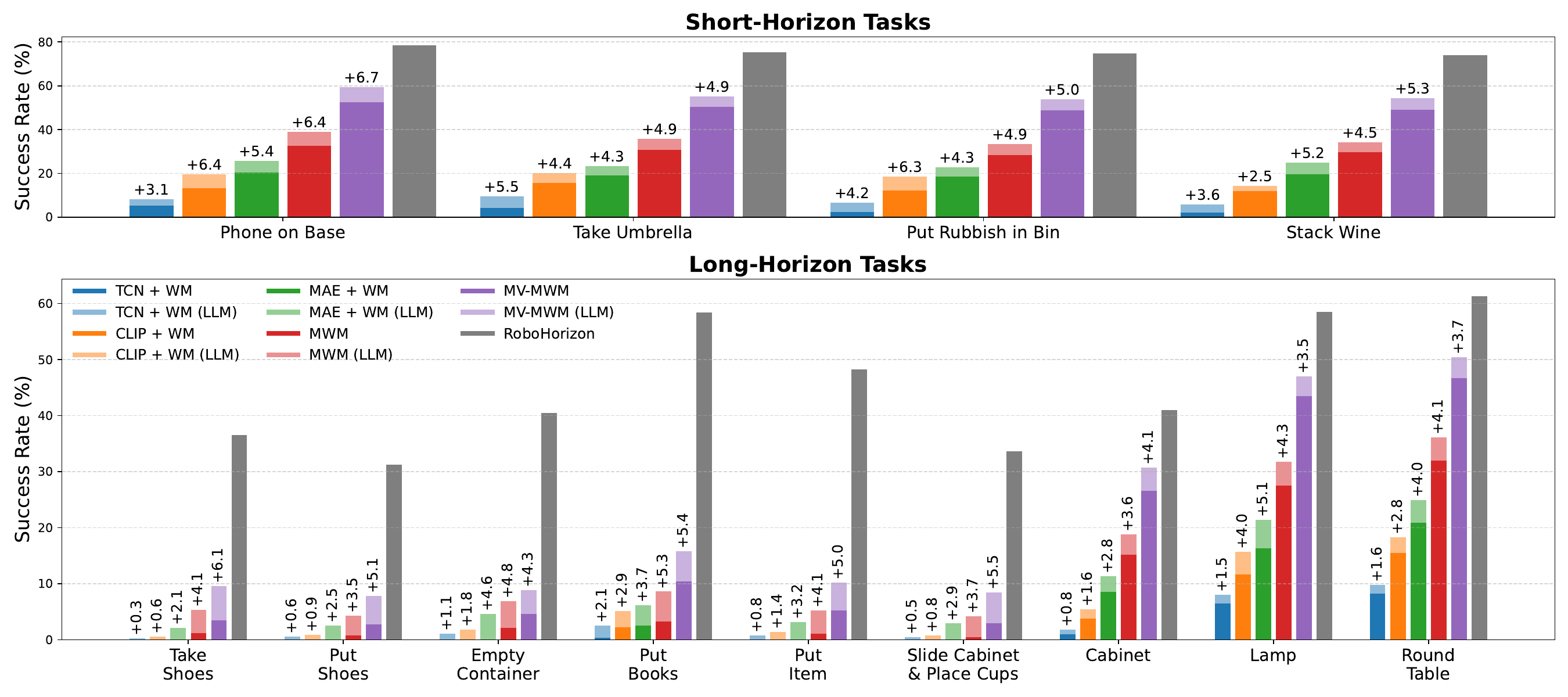}
    \vspace{-10pt}
    \caption{SPA-driven baselines with LLM-generated dense rewards vs. RoboHorizon.}
    \label{fig:spa_llm}
    \vspace{-0.6cm}
\end{figure*}

\paragraph{SPA-driven baselines with LLM-generated stepwise rewards vs. RoboHorizon}
To compare the performance of SPA-driven baselines with LLM-generated stepwise rewards but without staged rewards for each sub-task phase against RoboHorizon, we replace the manually defined reward system in SPA-driven baselines with the stepwise rewards generated during RoboHorizon’s Recognize phase using LLM assistance. Fig.~\ref{fig:spa_llm} shows the success rate comparison between RoboHorizon and five SPA-driven baselines enhanced with LLM-generated stepwise rewards (as indicated by ``LLM” in the legend) across \textcolor{ouryellow}{4 short-horizon tasks} and \textcolor{ourpurple}{9 long-horizon tasks}. It also illustrates the performance improvement of SPA-driven baselines with LLM-generated stepwise rewards compared to manually defined rewards.
The results show that while these baseline methods achieve some improvement in task success rates when using LLM-generated stepwise rewards, demonstrating the effectiveness of well-designed stepwise rewards aligned with motion planning in helping robots understand operational tasks, their performance gains in long-horizon tasks remain very limited without considering staged rewards for each sub-task phase. Furthermore, regardless of the task, they fail to surpass the full RoboHorizon framework driven by our proposed \textit{RSPA} pipeline.

This result clearly answers our second question: \textbf{By using LLM-generated stepwise rewards aligned with motion planning, SPA-driven baseline methods achieve certain performance improvements in both short-horizon and long-horizon tasks. However, due to their lack of consideration for staged rewards in each sub-task phase and their limited ability to learn complex long-horizon task representations, RoboHorizon still maintains a significant performance advantage.} This finding further highlights the importance of the LLM-assisted reward generation mechanism and key-horizon multi-view representation learning modules in the RSPA-driven RoboHorizon framework. These modules not only enhance the robot's ability to interpret task instructions but also improve its perception of tasks and target objects in complex long-horizon scenarios.

\begin{figure}
    \centering
    \includegraphics[width=0.45\textwidth]{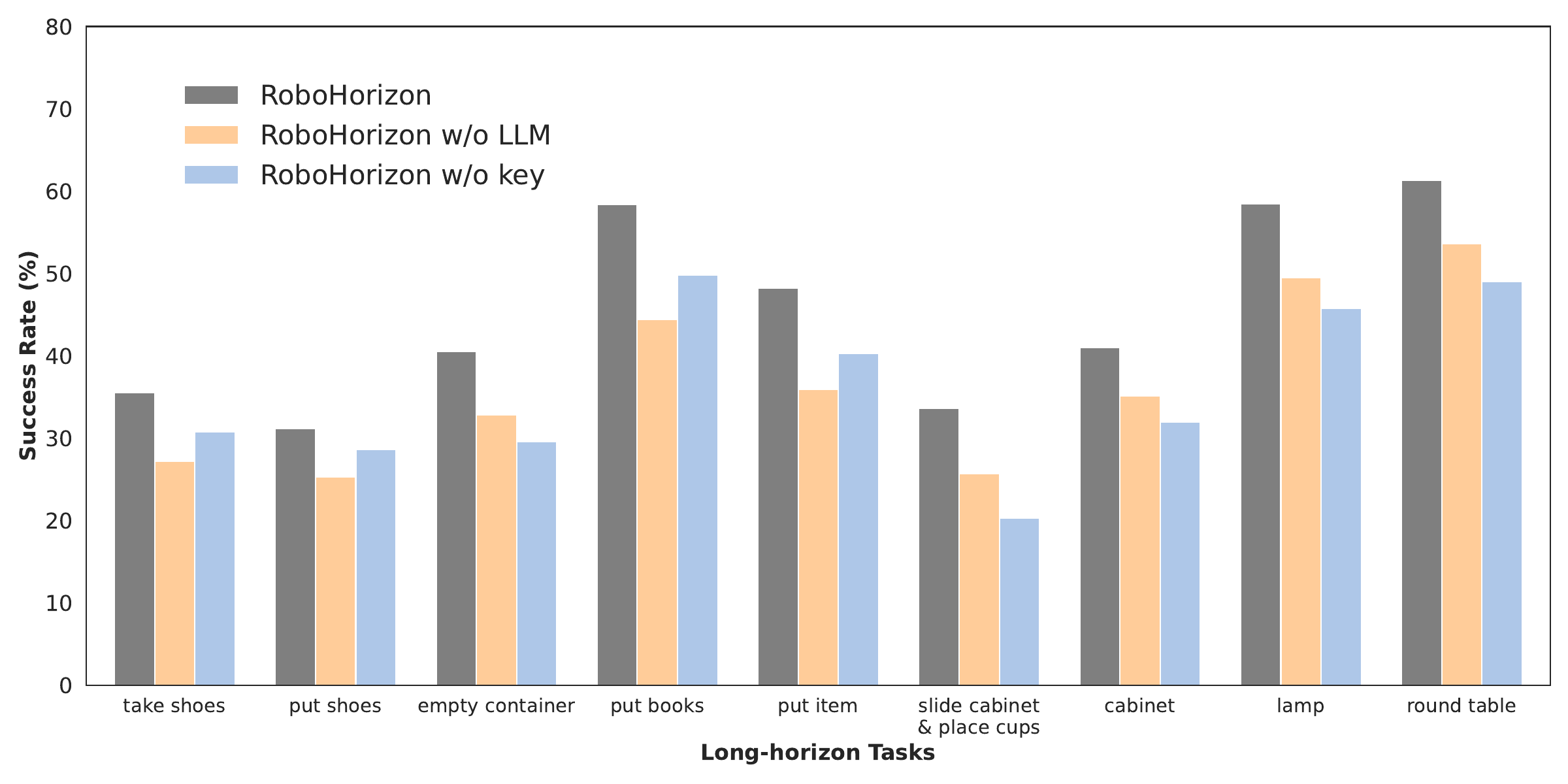}
    \vspace{-10pt}
    \caption{Ablation of key designs in RoboHorizon across 9 long-horizon tasks.}
    \label{fig:ablation}
    \vspace{-0.5cm}
\end{figure}

\paragraph{Ablation Study of RoboHorizon}
To evaluate the impact of RoboHorizon’s two key designs, LLM-assisted reward generation (recognition) and key-horizon multi-view representation learning (perception), on performance, we design two comparison methods: RoboHorizon w/o LLM, which removes LLM-assisted reward generation and relies on manually designed reward structures, and RoboHorizon w/o key, which omits key-horizon multi-view representation learning and uses the MV-MWM method for world model construction.
Fig.~\ref{fig:ablation} shows the ablation study results across nine long-horizon tasks. The results demonstrate that both LLM-assisted reward generation and key-horizon multi-view representation learning are critical to RoboHorizon’s performance, with each excelling in different scenarios. In tasks such as taking shoes out of a box, putting shoes in a box, placing books on a bookshelf, and putting items in a drawer, where objects are dispersed and representation learning is easier, RoboHorizon w/o key performs better. In contrast, in tasks like emptying a container, sliding a cabinet open and placing cups, and installing a cabinet, lamp, and round table, where objects are densely packed with more distractions, RoboHorizon w/o LLM performs better.

These findings clearly answer the third key question: \textbf{Both LLM-assisted reward generation and key-horizon multi-view representation learning are indispensable for RoboHorizon’s overall performance. In tasks with dispersed objects, LLM-assisted reward generation plays a more significant role, whereas in tasks with densely distributed objects, key-horizon multi-view representation learning is more crucial.}

\section{Conclusion and Discussion}
In this paper, we propose a novel \textit{Recognize-Sense-Plan-Act (RSPA)} pipeline for long-horizon manipulation tasks. The \textit{RSPA} pipeline decomposes complex long-horizon tasks into four stages: recognizing tasks, sensing the environment, planning actions, and executing them effectively. Based on this pipeline, we develop \textbf{RoboHorizon}, an LLM-assisted multi-view world model.
RoboHorizon uses its LLM-assisted reward generation module for accurate task recognition and its key-horizon multi-view representation learning module for comprehensive environment perception. These components enable RoboHorizon to build a robust world model that supports stable task planning and effective action execution through reinforcement learning algorithms.
Experiments on RLBench and FurnitureBench show that RoboHorizon significantly outperforms state-of-the-art baselines in long-horizon tasks. 
Future work will focus on two key areas: enhancing the LLM-assisted reward generation in RoboHorizon by incorporating human feedback to create a closed-loop process that strengthens the framework's adaptability to multiple tasks, and applying the RSPA pipeline and RoboHorizon model to real-world long-horizon robotic manipulation tasks to improve the framework's sim-to-real transfer capabilities.

\bibliographystyle{named}
\bibliography{ijcai25}

\clearpage
\appendix

\onecolumn

\section{Technical Appendix}

\begin{figure*}[t]  % Attempt to place the figure at the top
\centering
\includegraphics[width=0.95\textwidth]{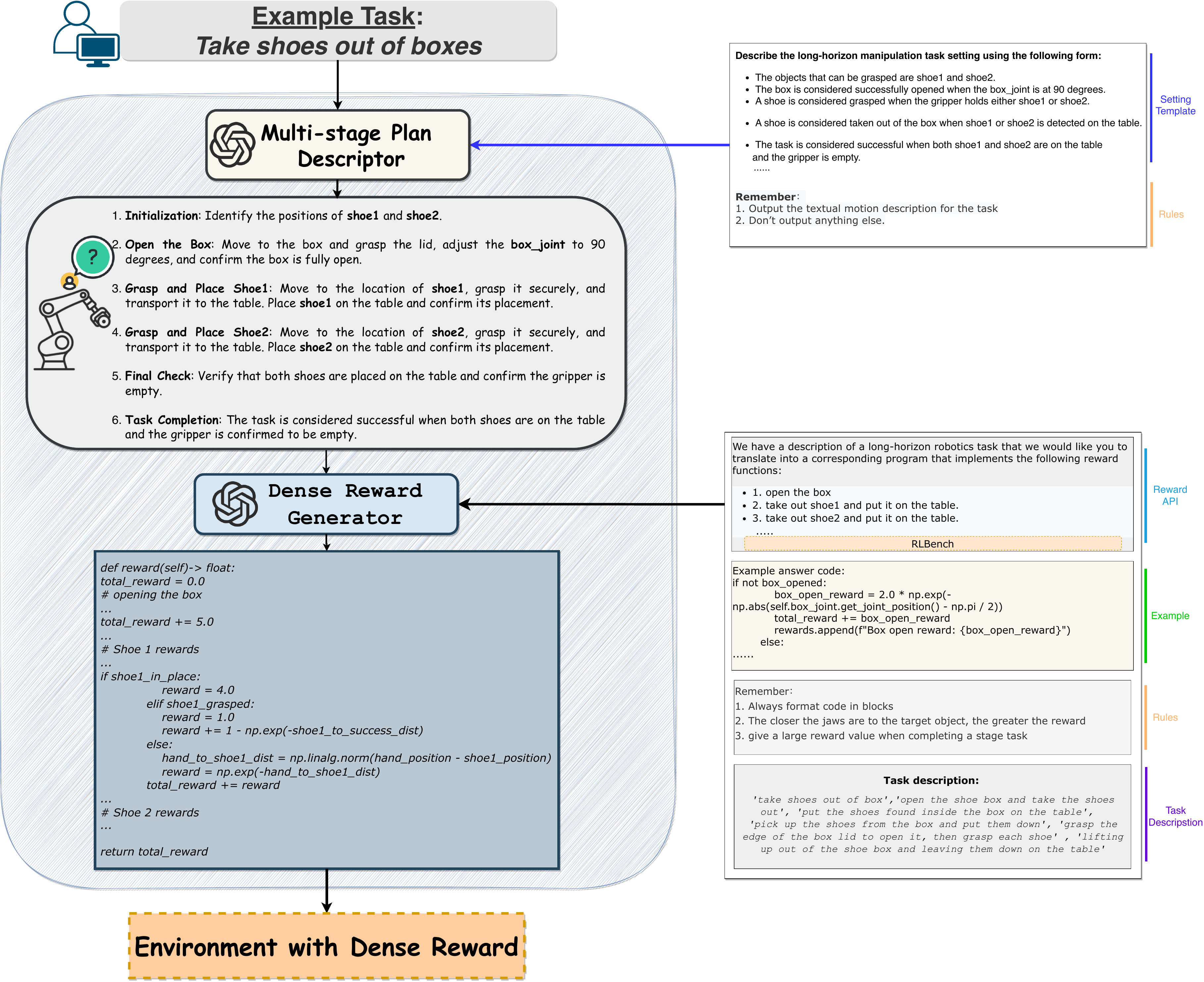} 
\caption{Detailed data flow of the LLM-assisted Reward Generation, the \textit{take shoes out of box} task as the illustration example.}
\label{reward}
\end{figure*}

In this technical appendix, we first provide a detailed example illustration of the data flow in the LLM-assisted Reward Generation (the \textbf{\textcolor{myorange}{Recognize}} part) and the two-stage (\textbf{Multi-stage Plan Descriptor} and \textbf{Dense Reward Generator}) prompts examples. 
Then, we present a detailed introduction to the RLBench and FurnitureBench benchmark used in the experiments, along with the related tasks.

\subsection{Detailed LLM-Assisted Reward Generation}

As shown in Fig.~\ref{reward}, the LLM-assisted Reward Generation process is divided into two stages. We use the long-horizon task \textit{``take shoes out of the box"} from RLBench as a demonstration to illustrate the two stages.

In Stage 1, we employ a pre-trained LLM as the \textbf{Multi-stage Plan Descriptor} that interprets and expands user input into detailed language descriptions of the required robot motions using predefined templates. To enable the multi-stage plan descriptor to generate a coherent structure for long-horizon tasks, we create a prompt template that outlines the current robot task setting. This leverages the pre-trained LLM's internal knowledge of motion planning to produce detailed motion descriptions. 

In Stage 2, we introduce another LLM as the \textbf{Dense Reward Generator}, which translates motion descriptions into corresponding dense reward structures. Specifically, for each sub-task stage, the model generates smaller \textbf{stepwise rewards} (e.g., rewards for each action during the grasping of shoe1) and larger \textbf{staged rewards} (e.g., rewards for successfully grasping shoe1). This process is treated as a coding task, leveraging the pre-trained LLM's deep understanding of code and its structure. The Dense Reward Generator is guided by four types of prompts to generate reward code: i) task stage descriptions based on the task environment interface, ii) examples of expected outputs from the reward generator, iii) constraints and rules for the reward encoder, and iv) specific task descriptions.

We provide a detailed two-stage data flow for the \textit{``take shoes out of the box"} task, along with the related prompts, the plans generated in Stage 1 and the reward codes generated in Stage 2.

% First tcolorbox with no space before or after
\begin{tcolorbox}[colback=black!10, colframe=gray, title=Data Flow for Stage 1, width=\textwidth, sharp corners, boxrule=0.5mm, toptitle=1mm, bottomtitle=1mm, coltitle=black, before skip=5pt, after skip=5pt, ]
\small
\textbf{Prompt:} Describe the long-horizon manipulation task \textit{``take shoes out of the box"} using the following format: 
\begin{itemize}[left=0pt]
    \item Graspable objects are shoe1 and shoe2.
    \item The box is considered successfully opened when the box\_joint is at 90 degrees.
    \item A shoe is considered grasped when the gripper holds either shoe1 or shoe2.
    \item A shoe is considered out of the box when it is detected on the table.
    \item The task is considered successful when both shoe1 and shoe2 are placed on the table.
\end{itemize}
Remember to:
\begin{itemize}[left=0pt]
    \item Provide only the motion description for the task.
    \item Exclude any other information.
\end{itemize}

\vspace{0.01cm} 

\small \textbf{Plan:} \textbf{[Motion Description for the Task]}
\begin{enumerate}[left=0pt]
    \item \textbf{Initialize:} Identify shoe1 and shoe2.
    \item \textbf{Open the Box:} Move towards the box and adjust the box\_joint to 90 degrees. Confirm the box is open.
    \item \textbf{Grasp and Place Shoe1:} Move towards shoe1, grasp it, move to the table, and place shoe1 on the table. Confirm shoe1 is on the table.
    \item \textbf{Grasp and Place Shoe2:} Move towards shoe2, grasp it, move to the table, and place shoe2 on the table. Confirm shoe2 is on the table.
    \item \textbf{Final Check:} Ensure both shoes are on the table and the gripper is empty.
    \item \textbf{Task Completion:} The task is successful when both shoes are placed on the table and the gripper is empty.
\end{enumerate}
\end{tcolorbox}

\vskip -0.2cm  % Remove space between the boxes

% Second tcolorbox
\begin{tcolorbox}[colback=black!10, colframe=gray, title=Data Flow for Stage 2, width=\textwidth, sharp corners, boxrule=0.5mm, toptitle=1mm, bottomtitle=1mm, coltitle=black, before skip=0pt, after skip=0pt]
\tiny
\textbf{Prompt:} We have a description of a long-horizon robotics task \textit{``take shoes out of the box"} that we would like you to translate into a corresponding code that implements the following reward functions:
\begin{itemize}[left=0pt]
    \item Open the Box: Move towards the box and adjust the box joint to 90 degrees. Confirm the box is open.
    \item Grasp and Place Shoe1: Move towards shoe1, grasp it, move to the table, and place shoe1 on the table. Confirm shoe1 is on the table.
    \item Grasp and Place Shoe2: Move towards shoe2, grasp it, move to the table, and place shoe2 on the table. Confirm shoe2 is on the table.
    \item Final Check: Ensure both shoes are on the table and the gripper is empty.
    \item Task Completion: The task is successful when both shoes are placed on the table and the gripper is empty.
\end{itemize}

\vspace{0.01cm}

\textbf{Example answer code:}
\begin{lstlisting}[breaklines=true, basicstyle=\ttfamily\tiny, frame=single]
if not box_opened:
    box_open_reward = 2.0 * np.exp(-np.abs(self.box_joint.get_joint_position() - np.pi / 2))
    total_reward += box_open_reward
    rewards.append(f"Box open reward: {box_open_reward}")
else:
    total_reward += 5.0  # Box opened reward
    rewards.append("Box opened reward: 5.0")
\end{lstlisting}

\vspace{0.01cm}

\textbf{Remember:}
\begin{itemize}[left=0pt]
    \item Always format code in blocks
    \item The closer the jaws are to the target object, the greater the reward
    \item Give a large reward value when completing a stage task
\end{itemize}

\vspace{0.01cm}

\textbf{Task description:} \\
'take shoes out of box', 'open the shoe box and take the shoes out', 'put the shoes found inside the box on the table',  
'pick up the shoes from the box and put them down', 'grasp the edge of the box lid to open it, then grasp each shoe', 'lift them up out of the shoe box and leave them down on the table'

\vspace{0.01cm}

\textbf{Generated Rewards:}
\begin{lstlisting}[breaklines=true, basicstyle=\ttfamily\tiny, frame=single]
def reward(self) -> float:
    rewards = []
    total_reward = 0.0

    box_opened = self._opened_cond_box.condition_met()[0]

    # Reward for opening the box
    if not box_opened:
        box_open_reward = 2.0 * np.exp(-np.abs(self.box_joint.get_joint_position() - np.pi / 2))
        total_reward += box_open_reward
        rewards.append(f"Box open reward: {box_open_reward}")
    else:
        total_reward += 5.0  # Box opened reward
        rewards.append("Box opened reward: 5.0")

    # Shoe 1 rewards
    shoe1_grasped = self._grasped_cond_shoe1.condition_met()[0]
    shoe1_in_place = self._detected_cond_shoe1.condition_met()[0]
    shoe1_position = self.shoe1.get_position()
    hand_position = self.robot.arm.get_tip().get_position()
    shoe1_to_success_dist = np.linalg.norm(shoe1_position - self.success_sensor.get_position())

    if shoe1_in_place:
        reward = 4.0
        rewards.append(f"Shoe1 in place reward: 4.0")
    elif shoe1_grasped:
        reward = 1.0
        rewards.append(f"Shoe1 grasped reward: 1.0")
        reward += 1 - np.exp(-shoe1_to_success_dist)
        rewards.append(f"Shoe1 moving towards target reward: {1 - np.exp(-shoe1_to_success_dist)}")
    else:
        hand_to_shoe1_dist = np.linalg.norm(hand_position - shoe1_position)
        reward = np.exp(-hand_to_shoe1_dist)
        rewards.append(f"Shoe1 approach reward: {np.exp(-hand_to_shoe1_dist)}")

    total_reward += reward

    # Shoe 2 rewards
    shoe2_grasped = self._grasped_cond_shoe2.condition_met()[0]
    shoe2_in_place = self._detected_cond_shoe2.condition_met()[0]
    shoe2_position = self.shoe2.get_position()
    shoe2_to_success_dist = np.linalg.norm(shoe2_position - self.success_sensor.get_position())

    if shoe2_in_place:
        reward = 4.0
        rewards.append(f"Shoe2 in place reward: 4.0")
    elif shoe2_grasped:
        reward = 1.0
        rewards.append(f"Shoe2 grasped reward: 1.0")
        reward += 1 - np.exp(-shoe2_to_success_dist)
        rewards.append(f"Shoe2 moving towards target reward: {1 - np.exp(-shoe2_to_success_dist)}")
    else:
        hand_to_shoe2_dist = np.linalg.norm(hand_position - shoe2_position)
        reward = np.exp(-hand_to_shoe2_dist)
        rewards.append(f"Shoe2 approach reward: {np.exp(-hand_to_shoe2_dist)}")

    total_reward += reward

    return total_reward
\end{lstlisting}
\end{tcolorbox}

\subsection{Experimental Details}
\paragraph{Simulation environment} We use the RLBench~\cite{james2020rlbench} and FurnitureBench~\cite{heo2023furniturebench} simulator. In the RLBench environment, we conduct experiments using a 7-DoF Franka Panda robot arm equipped with a parallel gripper on 4 short-horizon and 6 long-horizon visual manipulation tasks. In the FurnitureBench environment, we perform experiments with the same robot configuration on 3 long-horizon furniture assembly tasks.

\paragraph{Data collection}  To achieve key-horizon multi-view representation learning and policy learning that combines reinforcement learning with behavior cloning, we first collect expert data across both types of simulation tasks. For demonstration data collection in RLBench, we double the maximum velocity of the Franka Panda robot arm in PyRep~\cite{james2019pyrep}, which reduces the duration of the demonstrations without significantly compromising their quality. For each short-horizon task, we use RLBench's dataset generator to collect 50 demonstration trajectories per camera view, and for each long-horizon task, we collect 100 demonstration trajectories per camera view. For data collection in the FurnitureBench tasks, we utilize the automated furniture assembly scripts provided by the platform to automate the data collection process. Similarly, for each long-horizon furniture assembly task, we collect 100 demonstration trajectories per camera view.

\paragraph{Implementation} Our implementation is built on the official MV-MWM~\cite{seo2023multi} framework, and unless otherwise specified, the implementation details remain the same. To expedite training and mitigate the bottleneck caused by a slow simulator, we run 8 parallel simulators. Our autoencoder is composed of an 8-layer ViT encoder and a 6-layer ViT decoder, with an embedding dimension set to 256. We maintain a consistent set of hyperparameters across all experiments.

\paragraph{Computing hardware} For all RLBench experiments, we use a single NVIDIA GeForce RTX 4090 GPU with 24GB VRAM and it takes 12 hours for training MV-RoboWM and 16 hours for training MV-MWM.

\begin{figure}[h]
\centering
%\framebox[4.0in]{$\;$}
\includegraphics[width=0.85\textwidth]{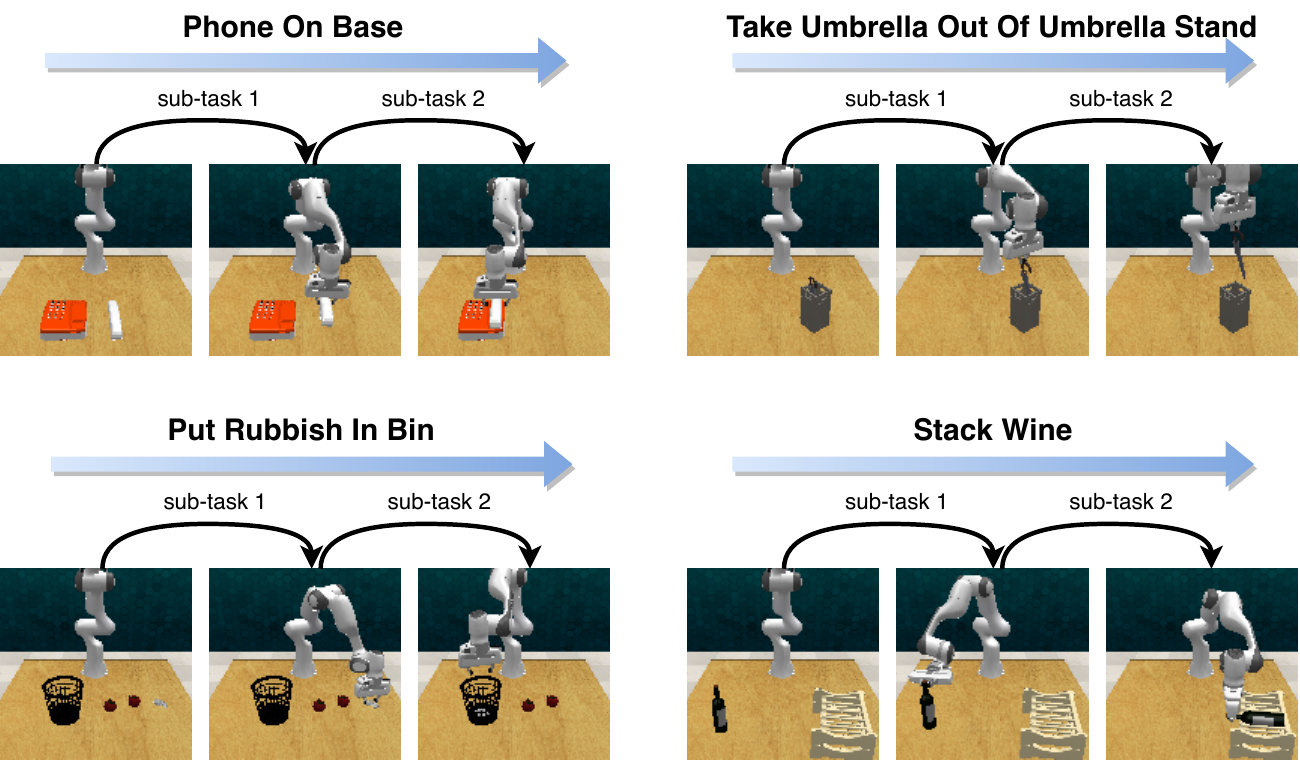} 
\caption{Visualization of the 4 short-horizon RLBench tasks in the experiment.}
\label{short_rlbenchtasks}
\end{figure}

\begin{figure*}[t!]
\centering
%\framebox[4.0in]{$\;$}
\includegraphics[width=0.73\textwidth]{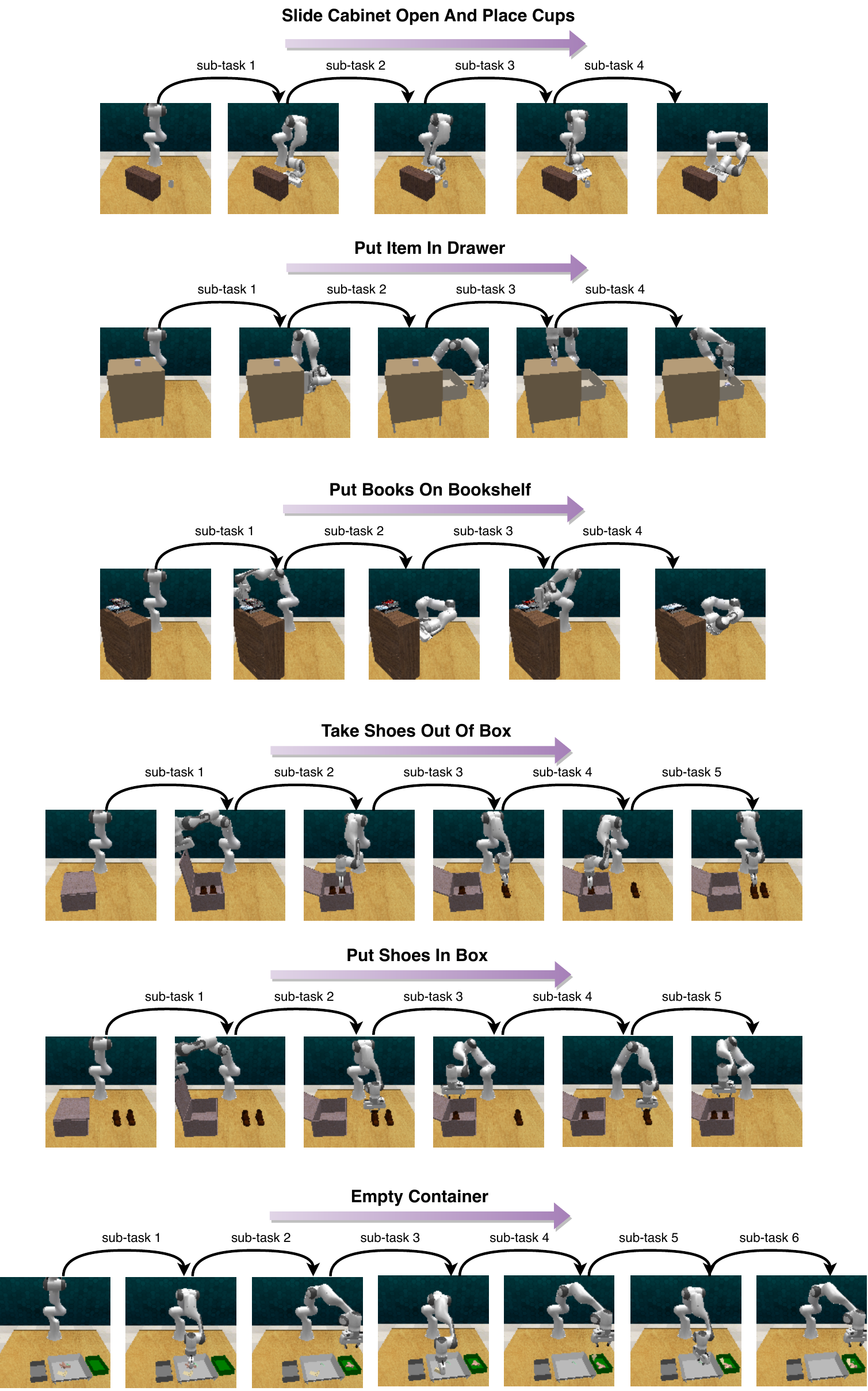} 
\caption{Visualization of the 6 long-horizon RLBench tasks in the experiment.}
\label{long_rlbenchtasks}
\end{figure*}

\subsection{RLBench Tasks} 
We select 10 tasks from the 100 available in RLBench for our simulation experiments, including 4 short-horizon tasks (as shown in Fig.~\ref{short_rlbenchtasks}) and 6 long-horizon tasks (as shown in Fig.~\ref{long_rlbenchtasks}). To reduce training time with limited resources, we only use \texttt{variation0}. In the following sections, we describe each of these 10 tasks in detail, including any modifications made to the original codebase.

\subsubsection{Phone On Base}
\textbf{Task:} grasp the phone and put it on the base.
\newline\textbf{filename: } \texttt{phone\_on\_base.py}
\newline\textbf{Modified: } Rewards are defined for each step based on the LLM-assisted reward generation module.
\newline\textbf{Success Metric: } The phone is on the base.
\newline\textbf{Task Horizon:} 2.

\subsubsection{Take Umbrella Out Of Umbrella Stand}
\textbf{Task:} grasp the umbrella by its handle, lift it up and out of the stand.
\newline\textbf{filename: } \texttt{take\_umbrella\_out\_of\_umbrella\_stand.py}
\newline\textbf{Modified: } Rewards are defined for each step based on the LLM-assisted reward generation module.
\newline\textbf{Success Metric: } The umbrella is taken off the stand.
\newline\textbf{Task Horizon:} 2.

\subsubsection{Put Rubbish In Bin}
\textbf{Task:} pick up the rubbish and leave it in the trash can.
\newline\textbf{filename: } \texttt{put\_rubbish\_in\_bin.py}
\newline\textbf{Modified: } Rewards are defined for each step based on the LLM-assisted reward generation module.
\newline\textbf{Success Metric: } The rubbish is thrown in the bin.
\newline\textbf{Task Horizon:} 2.

\subsubsection{Stack Wine}
\textbf{Task:} place the wine bottle on the wine rack.
\newline\textbf{filename: } \texttt{stack\_wine.py}
\newline\textbf{Modified: } Rewards are defined for each step based on the LLM-assisted reward generation module.
\newline\textbf{Success Metric: } The bottle is on the wine rack.
\newline\textbf{Task Horizon:} 2.

\subsubsection{Take Shoes Out Of Box}
\textbf{Task:} grasp the edge of the box lid to open it, then grasp each shoe, lifting up out of the shoe box and leaving them down on the table.
\newline\textbf{filename: } \texttt{take\_shoes\_out\_of\_box.py}
\newline\textbf{Modified: } Rewards are defined for each step based on the LLM-assisted reward generation module.
\newline\textbf{Success Metric: } Both shoes are placed on the table.
\newline\textbf{Task Horizon:} 5.

\subsubsection{Put Shoes In Box}
\textbf{Task:} open the box lid and put the shoes inside.
\newline\textbf{filename: } \texttt{put\_shoes\_in\_box.py}
\newline\textbf{Modified: } Rewards are defined for each step based on the LLM-assisted reward generation module.
\newline\textbf{Success Metric: } Both shoes are placed in the box.
\newline\textbf{Task Horizon:} 5.

\subsubsection{Empty Container}
\textbf{Task:} move all objects from the large container and drop them into the smaller red one.
\newline\textbf{filename: } \texttt{empty\_container.py}
\newline\textbf{Modified: } Rewards are defined for each step based on the LLM-assisted reward generation module.
\newline\textbf{Success Metric: } All objects in the large container are placed in the small red container.
\newline\textbf{Task Horizon:} 6.

\subsubsection{Put Books On Bookshelf}
\textbf{Task:} pick up 2 books and place them on the top shelf.
\newline\textbf{filename: } \texttt{put\_books\_on\_bookshelf.py}
\newline\textbf{Modified: } Rewards are defined for each step based on the LLM-assisted reward generation module.
\newline\textbf{Success Metric: } All the books are placed on top of the shelf.
\newline\textbf{Task Horizon:} 4.

\subsubsection{Put Item In Drawer}
\textbf{Task:} open the middle drawer and place the block inside of it.
\newline\textbf{filename: } \texttt{put\_item\_in\_drawer.py}
\newline\textbf{Modified: } Rewards are defined for each step based on the LLM-assisted reward generation module.
\newline\textbf{Success Metric: } The block is placed in the middle drawer.
\newline\textbf{Task Horizon:} 4.

\subsubsection{Slide Cabinet Open And Place Cups}
\textbf{Task:} grasping the left handle, open the cabinet, then pick up the cup and set it down inside the cabinet.
\newline\textbf{filename: } \texttt{slide\_cabinet\_open\_and\_place\_cups.py}
\newline\textbf{Modified: } Rewards are defined for each step based on the LLM-assisted reward generation module.
\newline\textbf{Success Metric: } The cup is in the left cabinet.
\newline\textbf{Task Horizon:} 5.

\subsection{FurnitureBench Tasks} 
\begin{figure*}[t!]
\centering
%\framebox[4.0in]{$\;$}
\includegraphics[width=\textwidth]{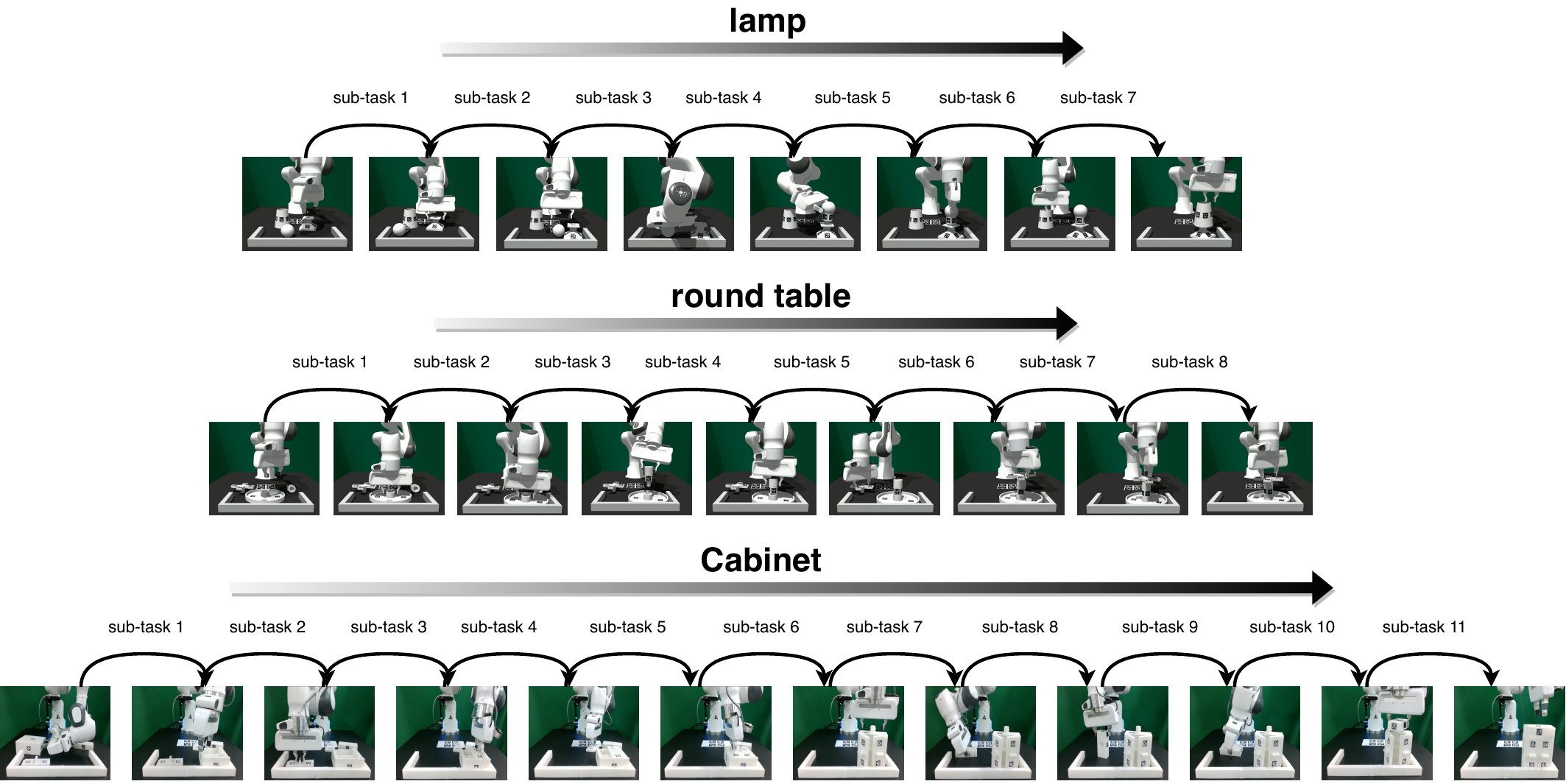} 
\caption{Visualization of the 3 long-horizon FurnitureBench tasks in the experiment, where the assembly of the cabinet is illustrated as the process with a real robot.}
\label{furnituretasks}
\end{figure*}

We select 3 furniture assembly tasks from the 9 available task in FurnitureBench for our experiments, as shown in Fig.~\ref{furnituretasks}). In the following parts, we describe each of these 3 tasks in detail, including any modifications made to the original codebase.

\subsubsection{Lamp}
\textbf{Task:} The robot needs to screw in a light bulb and then place a lamp hood on top of the bulb. The robot should perform
sophisticated grasping since the bulb can be easily slipped when grasping and screwing due to the rounded shape
\newline\textbf{Modified: } Rewards are defined for each step based on the LLM-assisted reward generation module.
\newline\textbf{Success Metric: } Three pieces of furniture are assembled into a lamp.
\newline\textbf{Task Horizon:} 7.

\subsubsection{Round Table}
\textbf{Task:} The robot should assemble one rounded tabletop, rounded leg, and cross-shaped table base. The robot
should handle an easily movable round leg and cross-shaped table base, which requires finding a careful grasping point.
\newline\textbf{Modified: } Rewards are defined for each step based on the LLM-assisted reward generation module.
\newline\textbf{Success Metric: } Three pieces of furniture are assembled into an round table.
\newline\textbf{Task Horizon:} 8.

\subsubsection{Cabinet}
\textbf{Task:} The robot must first insert two doors into the poles on each side of the body. Next, it must lock the top, so the
doors do not slide out. This task requires careful control to align the doors and slide into the pole. Moreover, diverse skills
such as flipping the cabinet body, and screwing the top are also needed to accomplish the task
\newline\textbf{Modified: } Rewards are defined for each step based on the LLM-assisted reward generation module.
\newline\textbf{Success Metric: } Four pieces of furniture are assembled into a cabinet.
\newline\textbf{Task Horizon:} 11.

\end{document}